\documentclass[11pt]{article}

\usepackage[margin=1in]{geometry}
\usepackage[round]{natbib}

\usepackage{amsthm,amsmath,amssymb,amsfonts}
\usepackage{tikz}
\usepackage{pgfplots}
\pgfplotsset{compat=1.18}
\usetikzlibrary{positioning, arrows.meta, fit, backgrounds, shapes.geometric, calc}
\usepackage{array}
\usepackage{graphicx}
\graphicspath{{Figs/}}
\usepackage{subcaption}
\usepackage{booktabs}
\usepackage{xcolor}
\usepackage{enumitem}
\usepackage{xspace}

\theoremstyle{tight}
\newtheorem{theorem}{Theorem}
\newtheorem{lemma}[theorem]{Lemma}
\newtheorem{corollary}[theorem]{Corollary}

\newcommand{\btheta}{{\boldsymbol\theta}}
\newcommand{\Ein}{E_{\mathrm{in}}}
\newcommand{\Eout}{E_{\mathrm{out}}}

\newcommand{\smallmath}[1]{%
  \par
  \vspace{-\parskip}%
  \begingroup
    \small
    \setlength{\abovedisplayskip}{0pt}%
    \setlength{\abovedisplayshortskip}{0pt}%
    \setlength{\belowdisplayskip}{4pt}%
    \setlength{\belowdisplayshortskip}{4pt}%
    #1%
  \endgroup
}

\title{Asymptotic Behavior of Transformers}

\makeatletter
\newcommand{\fnsym}[1]{%
  \ensuremath{^{\@fnsymbol{#1}}}
}
\makeatother
\author{
  Katy Blumer\fnsym{1} \qquad Da-Cheng Juan\fnsym{2} \qquad Jon Kleinberg\fnsym{3} \\
   Ravi Kumar\fnsym{4} \qquad  Cyrus Rashtchian\fnsym{2}  \qquad Andrew Tomkins\fnsym{5} \\[1ex]
  Google Research
}
\date{}

\begin{document}

\maketitle
\footnotetext[1]{\texttt{katyblumer@gmail.com}}
\footnotetext[2]{\texttt{\{dacheng,cyroid\}@google.com}}
\footnotetext[3]{\texttt{kleinberg@cornell.edu}}
\footnotetext[4]{\texttt{ravi.k53@gmail.com}}
\footnotetext[5]{\texttt{atomkins@gmail.com}}

\begin{abstract}

In this paper we study the behavior of transformer networks generating text over long periods of time.  We show that even a three-token,
single-layer attention-only transformer can fail to converge.   We then study a rich class of transformers with arbitrary finite depth, layer normalization, MLPs, RoPE encodings, and an arbitrary mix of local and global layers. Every transformer in this class has a  softmax temperature threshold above which its empirical cache distribution converges to a unique limit, independent of initial conditions. We show further that, while each transformer has a temperature threshold, the supremum of this set is infinite, so there is no global minimal temperature that forces all transformers to converge.

\end{abstract}

\section{Introduction}

In this paper we study what happens
if an autoregressive transformer is allowed to generate for a very long time.  Most theoretical analyses of transformers fix the sequence length, or ask how performance changes when the
context window is extended~\citep{kazemnejad2023positional, huang2025formal,sun2023length}.  Here we ask a different question: \textsl{we fix the model and study the
generation process itself as the number of generated tokens tends to infinity}.

As a side note, while we think this question is mathematically natural, we are also motivated by the observation that modern models sometimes seem to ``lock in'' to a particular state~\citep{SharmaTKDABDHJK24, HuangICLR24}. Sometimes this state is annoyingly bad, such that even with great efforts the model cannot be convinced to let a particular approach or argument go. Other times the state is pleasingly good, perhaps encoding an effective understanding of a problem that represents a useful starting point for many further explorations---in this case we might wish to checkpoint the model in this good state, to use it as a resource. These observations led us to wonder, even without a human in the loop, whether a model might enter a state that continues to influence its behavior forever. This paper provides a partial answer.


\paragraph{Single-Layer Transformers.}
Our first goal is to understand this question in a simple setting: a single-layer attention-only transformer without positional
encodings, MLPs, or normalization~\citep{vaswani2017attention,kazemnejad2023positional,wang2024length}. In this model, different occurrences
of the same token are indistinguishable to attention, so the entire
history is summarized by the empirical token distribution $P_t$. As $t$ grows large, the change to the historical distribution in a single step goes to zero; hence, for fixed-size finite windows and sufficiently large $t$, we may view the historical state of the model as frozen.

The next token logits depend only on the current token and the history, so freezing the historical distribution at some $P$ causes the simplified transformer to behave like a finite-state
Markov chain with transition matrix $M_P$. The historical distribution will slowly drift towards the steady state of this Markov chain, $\pi^*(P)$, but as it does so, the Markov chain itself will evolve, resulting in a new drift direction. We show that the empirical
distribution $P_t$ tracks the ODE almost surely~\citep{benaim1999dynamics}
\[
\frac{dP}{dt}=\pi^*(P)-P.
\]
Thus the
long-run behavior of this model is described by a deterministic dynamical system, even though the
underlying token sequence remains random.

The softmax temperature $\tau$ enters this ODE in a particularly simple way. Differentiating
the output softmax contributes a factor $1/\tau$, and we use this to
show that, for this model, there is a finite
temperature above which the map $P\mapsto\pi^*(P)$ is a contraction.
In this regime the ODE has a unique globally attracting fixed point and
$P_t$ converges to it almost surely, independently of the initial
prompt; as $\tau\to\infty$, this fixed point approaches the uniform
distribution. At lower temperatures the ODE description remains valid,
but uniqueness of the limiting set is no longer guaranteed; in fact, even this simplified
model need not converge.

\paragraph{Multi-Layer Transformers.}
We next consider fixed-depth transformers,
allowing arbitrary mixtures of global and fixed-window local attention
layers~\citep{beltagy2020longformer,zaheer2020bigbird}, RoPE positional encodings~\citep{su2021roformer}, MLPs, residual connections, and
normalization. Here the empirical token distribution is no longer
sufficient to describe the history. The key/value vectors stored at
a position depend on the context in which that position was generated,
so two occurrences of the same token can contribute different entries
to the global KV cache.

Nevertheless, each historical position contributes only a
fixed-dimensional vector to this cache. We therefore represent the
unbounded global cache by the empirical distribution $\eta_t$ of these
vectors. The dimension of an individual cache vector depends on the
fixed architecture but not on the generation time. In this sense the entire historical
state remains fixed-dimensional at each historical position even though
the number of stored positions grows without bound.

RoPE position encodings introduce an additional state variable: the current rotation
phase. With the global
cache frozen at $\eta$, the finite local-attention windows imply that
only a fixed number $W$ of recent tokens is needed in addition to the
current RoPE phase $\omega$. The resulting process over $W$-grams is governed by a
sequence of phase-dependent transition matrices
$Q_{\eta,\omega}$.

Again with frozen history $\eta$, at any fixed positive temperature, every output token has a probability bounded
away from zero. Every $W$-gram can be reached from every
other in $W$ steps, and the effect of the initial $W$-gram decays
geometrically. This gives a unique
phase-dependent family $\pi_\eta(\omega)$ describing the asymptotic
distribution of the recent tokens. We show that this frozen-cache
description remains accurate over sufficiently long blocks that the local state at the start of the block is forgotten as the system moves towards the frozen-cache limiting behavior.

The RoPE phase itself does not converge, but the averages along its orbit do.
Averaging the phase-dependent local behavior therefore defines a map
$\Gamma(\eta)$: the average distribution over all RoPE phases of the next cache vector when
the current cache is frozen at $\eta$. Since the empirical cache changes
by only $1/(t+1)$ at each step, a natural time scale for state evolution is
\[
s_t=\sum_{j=1}^t\frac1j=\log t+O(1).
\]
On this scale we obtain the measure-valued ODE
\[
\frac{d\eta}{ds}=\Gamma(\eta)-\eta.
\]
We prove that the actual empirical cache process almost surely tracks
this ODE over every fixed interval of the $s$-time scale. Thus the
long-run behavior of the full multi-layer transformer is again reduced
to a deterministic dynamical system, now over probability distributions
on a fixed-dimensional cache space.

This description by itself does not imply convergence: the ODE may (and in fact does) have more complicated non-convergent trajectories. We therefore return to
the role of temperature. The feed-forward ordering of transformer
layers gives a useful triangular structure: the cache vector written by
a global layer can depend only on the caches of earlier global layers.
Using a weighted distance across layers, this triangular dependence can
be made arbitrarily small. The remaining feedback through token
sampling is suppressed by the $1/\tau$ derivative of the output
softmax. We use these two facts to show that every fixed transformer in
our model has a finite temperature threshold above which $\Gamma$ is a
contraction. Above this threshold the cache ODE has a unique globally
attracting fixed point, and the empirical cache converges to it almost
surely.

The threshold, however, cannot be chosen uniformly over all
transformers.  We give an explicit three-token example showing that
even a single-layer, NoPE, attention-only transformer with no MLP can
fail to converge almost surely.  Moreover, for every finite
temperature $\tau$, scaling its output embeddings by $\tau$ leaves the
next token probabilities unchanged.  Thus for every finite temperature
there is such a simplified transformer that fails to converge, even
though every fixed transformer converges once its temperature is
sufficiently large.

The order of quantifiers is important. For every fixed transformer,
sufficiently high temperature forces convergence; but for every fixed
finite temperature there is some transformer that fails to converge.
Hence there is no universal finite temperature that guarantees
convergence over the entire class.

Our results therefore give two complementary descriptions of
long-generation behavior. First, slowly changing empirical state turns
the stochastic generation process into a deterministic ODE on an
appropriate time scale. Second, sufficiently high temperature controls
the feedback in this ODE and forces convergence for each fixed model,
while no single temperature can do so uniformly for all models.

\subsection{Background}

Throughout the paper, $\Sigma$ denotes a finite vocabulary with
$|\Sigma|=V$, and $x_t\in\Sigma$ is the token generated at time $t$.
Let $\mathcal F_t$ denote the information contained in the generated
history $(x_1,\ldots,x_t)$, and let $\tau>0$ denote the \emph{sampling
temperature}.  We write $\Delta^{V-1}$ for the probability simplex.

For vectors, $\langle\cdot,\cdot\rangle$ and $\|\cdot\|$ denote the
Euclidean inner product and norm; for matrices, $\|\cdot\|$ denotes
the induced operator norm.  For distributions $\mu,\nu$,
 $\|\mu-\nu\|_{\rm tv} :=
\sup_A |\mu(A)-\nu(A)|$ 
denotes the \emph{total variation distance}, and $d_1(\mu,\nu) :=
\sup_{\operatorname{Lip}(\varphi)\leq1}
\left|
\int\varphi\,d\mu-\int\varphi\,d\nu
\right|$ 
denotes the \emph{$1$-Wasserstein distance}.
Two distances on distributions are \emph{equivalent} if they differ by at most constant factors.  
We use $\delta_{\cdot,\cdot}$ for
the Kronecker delta and $\delta_{z}$ for the unit 
point mass at $z$.

\paragraph{Stochastic Approximation.}
We repeatedly use the dynamical systems approach to stochastic
approximation of \citet{benaim1999dynamics}.  Informally speaking, for an update
of the form
\[
X_{t+1}=X_t+\gamma_t\bigl(F(X_t)+\text{noise}\bigr),
\]
where the step sizes $\gamma_t$ are
decreasing and satisfy 
$\sum_t\gamma_t=\infty$ and $\sum_t\gamma_t^2<\infty$, the process,
viewed on the accumulated time scale
$s_t=\sum_{j<t}\gamma_j$, tracks the ODE
$dX/ds=F(X)$ over every fixed interval of $s$-time.  Under the usual
boundedness and noise conditions, its limiting set is an internally
chain transitive (ICT)%
\footnote{
A non-empty compact set $K$ is
an \emph{internally chain transitive set (ICT)} of the ODE $\frac{dP}{dt} = f(P)$ if for every $P, P' \in K$, every 
$\epsilon > 0$, and every $T > 0$, there exists points $P = P_0, P_1, \ldots, P_m = P' \in K$ and times $t_0, \ldots, t_{m-1} \geq T$ such that $\| \phi_{t_i}(P_i) - P_{i+1} \| \leq \epsilon, \forall i$, where $\phi_t(P)$ is the solution of the ODE.  
} 
set of this ODE.  We use this framework in
Section~\ref{sec:single}
to obtain the ODE for the empirical token distribution, and
in Section~\ref{sec:multi}
to obtain the analogous ODE for the empirical cache
distribution.  Once the corresponding ODE has a unique globally
attracting fixed point, this tracking gives convergence of the
stochastic process.  We also use Bena\"im's result that stochastic
approximation does not converge to a linearly unstable equilibrium
under nondegenerate noise in 
Section~\ref{sec:nonconvergence}.

\section{Single-Layer Transformer}
\label{sec:single}

In this section we study a simple transformer model.
The model has a single layer with one attention head, and lacks positional embeddings (NoPE), layer normalization, and MLPs. Let $d$ be the \emph{hidden dimension}. The model is parameterized by fixed input and output embedding matrices $\Ein, \Eout \in \mathbb{R}^{V \times d}$, and standard attention projection matrices $W_Q, W_K, W_V \in \mathbb{R}^{d \times d}$.  

At step $t$, the state is given by the current token $x_t$ and the historical context. With NoPE, attention to each token of the history is independent of the token's location and so, the \emph{history} is captured by the empirical distribution $P_t \in \Delta^{V-1}$, where $P_t(v)$ is the frequency of token $v$ up to time $t$. 

For a query token $u$ and key/value token $w$, the unnormalized \emph{attention score} is $s_{u,w} = \frac{1}{\sqrt{d}} \langle (\Ein[u]W_Q), (\Ein[w]W_K) \rangle$. 
The attention weight that $u$ pays to $w$ over history $P_t$ is:
\begin{eqnarray}
\label{eq:attnweight}
    \alpha_{u,w}(P_t) & = & \frac{P_t(w) \exp(s_{u,w})}{Z_u(P_t)}, \mbox{ where } \nonumber \\
    Z_u(P_t) & = & \sum_{w=1}^V P_t(w) \exp(s_{u,w}). 
\end{eqnarray}

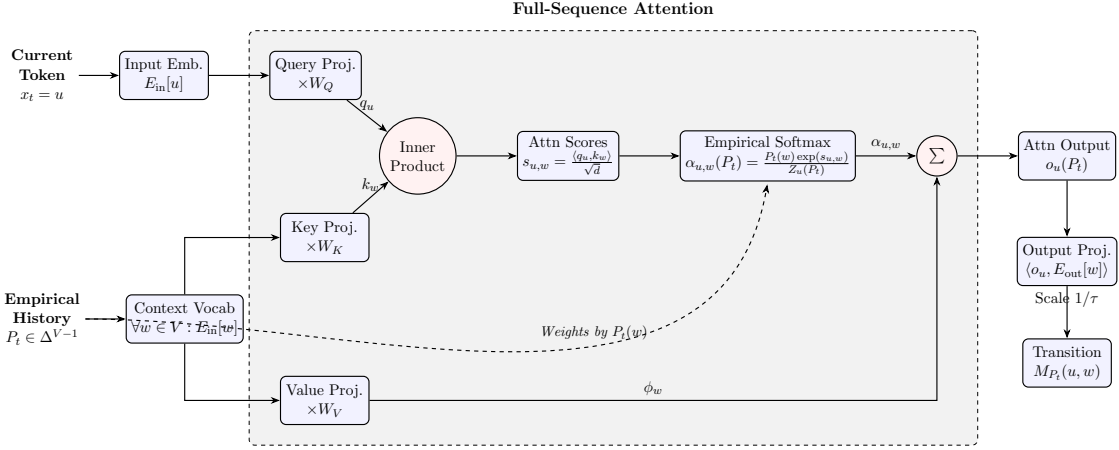
\begin{figure}[tb]
\centering
\resizebox{0.9\textwidth}{!}{
\begin{tikzpicture}[
    >=Stealth,
    node distance=1.5cm and 1.8cm,
    box/.style={draw, rectangle, rounded corners, align=center, minimum height=1.2cm, minimum width=2.2cm, fill=blue!5, thick},
    circ/.style={draw, circle, align=center, minimum size=1cm, fill=red!5, thick},
    txt/.style={align=center, font=\bfseries},
    thick
]

\node[txt] (xt) {Current \\Token\\$x_t = u$};
\node[txt, below=4.5cm of xt] (pt) {Empirical \\ History\\$P_t \in \Delta^{V-1}$};

\node[box, right=1cm of xt] (ein_u) {Input Emb.\\$\Ein[u]$};
\node[box, right=1cm of pt] (ein_w) {Context Vocab\\$\forall w \in V: \Ein[w]$};

\node[box, right=1.5cm of ein_u] (wq) {Query Proj.\\$\times W_Q$};
\node[box, above right=0.8cm and 0.9cm of ein_w] (wk) {Key Proj.\\$\times W_K$};
\node[box, below right=0.8cm and 0.9cm of ein_w] (wv) {Value Proj.\\$\times W_V$};

\node[circ, below right=1cm of wq] (dot) {Inner\\Product};
\node[box, right=1.5cm of dot] (score) {Attn Scores\\$s_{u,w} = \frac{\langle q_u, k_w \rangle}{\sqrt{d}}$};

\node[box, right=1.5cm of score] (alpha) {Empirical Softmax\\$\alpha_{u,w}(P_t) = \frac{P_t(w) \exp(s_{u,w})}{Z_u(P_t)}$};

\node[circ, right=1.5cm of alpha] (sum) {$\sum$};
\node[box, right=1.5cm of sum] (ou) {Attn Output\\$o_u(P_t)$};

\node[box, below=1.4cm of ou] (logits) {Output Proj.\\$\langle o_u, \Eout[w] \rangle$};
\node[box, below=1.3cm of logits] (transition) {Transition\\$M_{P_t}(u,w)$};

\draw[->] (xt) -- (ein_u);
\draw[->] (pt) -- (ein_w);

\draw[->] (ein_u) -- (wq);
\draw[->] (ein_w) |- (wk);
\draw[->] (ein_w) |- (wv);

\draw[->] (wq) -- node[above] {$q_u$} (dot);
\draw[->] (wk) -- node[above] {$k_w$} (dot);
\draw[->] (dot) -- (score);

\draw[->] (score) -- (alpha);

\draw[->, dashed, shorten >=2pt] (pt) to[out=0, in=260] node[pos=0.6, right, above, xshift=0.2cm, font=\small\itshape] {Weights by $P_t(w)$} (alpha.south);

\draw[->] (alpha) -- node[above] {$\alpha_{u,w}$} (sum);
\draw[->] (wv) -| node[near start, right, above] {$\phi_w$} (sum);

\draw[->] (sum) -- (ou);
\draw[->] (ou) -- (logits);
\draw[->] (logits) -- node[above] {Scale $1/\tau$} (transition);

\begin{scope}[on background layer]
    \node[fit=(wq)(wk)(wv)(dot)(score)(alpha)(sum), draw, dashed, fill=gray!10, rounded corners, inner sep=0.5cm] (attn_bg) {};
    \node[above=0.2cm of attn_bg.north, font=\bfseries\large] {Full-Sequence Attention};
\end{scope}

\end{tikzpicture}
}
\caption{A stripped-down model of a transformer with NoPE, no MLPs, and an unbounded context window parameterized by historical distribution $P_t$. \label{fig:transformer}}
\end{figure}

Let $\phi_w = \Ein[w]W_V$ denote the value vector for token $w$. The 
\emph{self-attention output} is $o_u(P_t) = \sum_{w=1}^V \alpha_{u,w}(P_t) \phi_w$. The probability of generating the next token $w$ given query $u$ and history $P_t$, sampled at temperature $\tau > 0$, defines the \emph{transition matrix} $M_{P_t} \in \mathbb{R}^{V \times V}$ 
(see Figure~\ref{fig:transformer}):
\[
    M_{P_t}(u,w) = \frac{\exp\left( \langle o_u(P_t), \Eout[w] \rangle / \tau \right)}{\sum_{w'=1}^V \exp\left( \langle o_u(P_t), \Eout[w'] \rangle / \tau \right)}.
\]

\subsection{Limiting Dynamics}\label{sec:mean-field-dynamics}

We first describe the asymptotic evolution of the empirical distribution $P_t$ by an ordinary differential equation (ODE).
Observe that $P_t$ 
updates iteratively as:
\[
    P_{t+1}(v) = P_t(v) + \frac{1}{t+1} \left( \mathbb{I}(x_{t+1} = v) - P_t(v) \right).
\]
To analyze the long-run behavior of $P_t$, we appeal to the theory of stochastic approximation.   To show that the discrete stochastic process tracks a deterministic
limiting ODE, we verify the standard conditions for stochastic
approximation~\citep{benaim1999dynamics}:

(i) \textit{Step-size sequence:} The update step size $\gamma_t := \frac{1}{t+1}$ satisfies the conditions: $\sum_t \gamma_t = \infty$ 
and $\sum_t \gamma_t^2 < \infty$.

(ii) \textit{Compact state space:}   $P_t$ lies in the simplex $\Delta^{V-1}$, which is compact.

(iii) \textit{Smoothness:}  The ODE has drift $\pi^*(P) - P$, where $\pi^*(P)$ is the stationary distribution of $M_P$. Since $\tau>0$ and the logits are bounded on the simplex, each entry in $M_P$ is bounded away from zero; thus, $M_P$ is uniformly ergodic. Since $M_P$ also depends smoothly on $P$, $\pi^*(P)$ is continuously differentiable
(and hence Lipschitz) on the simplex.

(iv) \textit{Markovian state noise and martingale difference:}  
Since the next token $x_{t+1}$ is sampled from row $x_t$ of $M_{P_t}$, its conditional expectation given $\mathcal{F}_t$ is $M_{P_t}(x_t, \cdot)$ rather than $\pi^*(P_t)$. We therefore decompose the update increment as:
\begin{align*}
    & \mathbb{I}(x_{t+1} = v) - P_t(v) = \underbrace{\big(\pi^*(P_t)(v) - P_t(v)\big)}_{\text{mean-field drift}} \\ 
    & + \underbrace{\big(M_{P_t}(x_t, v) - \pi^*(P_t)(v)\big)}_{\text{Markovian state noise}}  
    + \underbrace{\big(\mathbb{I}(x_{t+1} = v) - M_{P_t}(x_t, v)\big)}_{\text{martingale difference}}.
\end{align*}
%
\begin{lemma}
\label{lem:odeict}
For every fixed $\tau>0$, the empirical distribution $P_t$ tracks the ODE a.s.
\[
\frac{dP}{dt}=\pi^*(P)-P.
\]
Hence, the limiting set of $P_t$ is an ICT of this ODE a.s.  
\end{lemma}
\begin{proof}
Using (iv), the update is given by 
\[
P_{t+1}-P_t
=
\gamma_t\bigl(\pi^*(P_t)-P_t+\xi_{t+1}+\eta_{t+1}\bigr),
\qquad
\]
where
\[
\xi_{t+1}=M_{P_t}(x_t,\cdot)-\pi^*(P_t),
\enspace
\eta_{t+1}=I(x_{t+1}=\cdot)-M_{P_t}(x_t,\cdot).
\]
Note that $\eta_{t+1}$ is a bounded martingale difference, and so
$\sum_t \gamma_t\eta_{t+1}$ converges (a.s.) as
$\sum_t\gamma_t^2<\infty$ from (i).

For the Markovian term $\xi_{t+1}$, uniform ergodicity from (iii) implies that
there is a uniformly bounded function $g_P$, Lipschitz-continuous in $P$,
satisfying
\[
g_P(x)-\sum_y M_P(x,y) g_P(y)
=
M_P(x,\cdot)-\pi^*(P).
\]
Thus
\[
\xi_{t+1}
=
g_{P_t}(x_t)-g_{P_t}(x_{t+1})+\zeta_{t+1},
\]
where
\[
\zeta_{t+1}
=
g_{P_t}(x_{t+1})
-
\sum_y M_{P_t}(x_t,y) g_{P_t}(y),
\]
is a bounded martingale difference.
After multiplication by $\gamma_t$, the first two terms telescope up to a
boundary term, which vanishes, and errors of order
$|\gamma_t-\gamma_{t-1}|$ and
$\gamma_t\|P_t-P_{t-1}\|=O(\gamma_t^2)$, which are summable.
Also, $\sum_t\gamma_t\zeta_{t+1}$ converges (a.s.). Hence the accumulated
Markovian noise has vanishing tails.

Since $P\mapsto\pi^*(P)-P$ is Lipschitz, the proof follows from stochastic approximation~\citep{benaim1999dynamics}.
\end{proof}

%
%

\subsection{Jacobian}
In this section, we study the limiting behavior of our simple transformer.  In particular, we show that there is a  temperature threshold above which the limit of the empirical distribution $P_t$ is a unique fixed point.  

From Lemma~\ref{lem:odeict}, the transformer tracks the ODE in the limit, so we study now whether all solutions converge to the same limit, or whether different limiting sets are possible.  To do this, we analyze the Jacobian
$J\in\mathbb{R}^{V\times V}$ of the map $\pi^*(P)$. This Jacobian
characterizes how the steady state of the Markov chain moves in response
to a local change in the history. Its entries are
$J_{w,v}=\partial\pi^*(w)/\partial P(v)$.

Note that taking a partial derivative with respect to a single component $P(v)$ while holding all other components $P(j \neq v)$ constant  points in a direction that leaves the probability simplex (since the sum of the components will no longer equal 1). Since the formulas above extend smoothly
to a neighborhood of the simplex, we can use these derivatives to
assemble the full Jacobian in $\mathbb{R}^{V\times V}$.  Since $\sum_w \pi^*(w)=1$, only perturbations whose coordinates sum to zero
correspond to valid changes of the probability distribution. Restricting
the Jacobian to these perturbations gives the correct local behavior on the simplex.
\begin{lemma} \label{lem:jacobian}
It holds that
    \[
    \frac{\partial \pi^*}{\partial P(v)} = \frac{\partial \pi^*}{\partial P(v)} M_P + \pi^* \frac{\partial M_P}{\partial P(v)}.
\]
\end{lemma}
\begin{proof}
We trace the chain of derivatives with respect to $P(v)$:

(i) We differentiate the normalization term $Z_u$ in \eqref{eq:attnweight}:
\[
    \frac{\partial Z_u}{\partial P(v)} = \exp(s_{u,v}).
\]
Applying the quotient rule to  $\alpha_{u,k}$ in
\eqref{eq:attnweight}:
\begin{align*}
    \frac{\partial \alpha_{u,k}}{\partial P(v)} & = \frac{\delta_{k,v} \exp(s_{u,k}) Z_u - P(k)\exp(s_{u,k})\exp(s_{u,v})}{Z_u^2} \\
    & = \frac{\exp(s_{u,v})}{Z_u} (\delta_{k,v} - \alpha_{u,k}).
\end{align*}
Summing over all $k$ to find the shift in the attention output $o_u$:
\begin{equation}
\label{eq:attnoutputpd}
    \frac{\partial o_u}{\partial P(v)} = \sum_{k=1}^V \frac{\partial \alpha_{u,k}}{\partial P(v)} \phi_k = \frac{\exp(s_{u,v})}{Z_u} (\phi_v - o_u).
\end{equation}

(ii) Let $z_{u,w} = \langle o_u, \Eout[w] \rangle$ be the logit for token $w$. Its derivative is linear: $\frac{\partial z_{u,w}}{\partial P(v)} = \left\langle \frac{\partial o_u}{\partial P(v)}, \Eout[w] \right\rangle$.
The transition probability $M_P(u,w)$ is a softmax over these logits scaled by $\tau$. Applying the standard softmax derivative, the chain rule,
and using \eqref{eq:attnoutputpd}, 
\begin{align*}
& \frac{\partial M_P(u, w)}{\partial P(v)} = \sum_{y \in V} \frac{\partial M_P(u, w)}{\partial z_{u,y}} \frac{\partial z_{u,y}}{\partial P(v)} \nonumber \\
&= \sum_{y \in V} \frac{1}{\tau} M_P(u, w) (\delta_{w,y} - M_P(u, y)) \left\langle
\frac{\partial o_u}{\partial P(v)},E_{\rm out}[y]
\right\rangle \nonumber \\
    &= \frac{1}{\tau} M_P(u, w) \left\langle
\frac{\partial o_u}{\partial P(v)},E_{\rm out}[w]
\right\rangle \\
    & \qquad - \sum_{y \in V} M_P(u, y) \left\langle
\frac{\partial o_u}{\partial P(v)},E_{\rm out}[y]
\right\rangle.
\end{align*}
Let $\bar{e}_u = \sum_{y \in V} M_P(u, y) \Eout[y]$ be the expected output embedding over the entire vocabulary for query $u$.  Rewriting, we obtain
\smallmath{
\begin{equation}
\label{eq:partialM}
    \frac{\partial M_P(u, w)}{\partial P(v)} = \frac{1}{\tau} M_P(u, w) \frac{\exp(s_{u,v})}{Z_u} \left\langle
\phi_v-o_u,\,
E_{\rm out}[w]-\bar e_u
\right\rangle.
\end{equation}
}
(iii) Finally, we derive the Jacobian of the stationary distribution $J_{w,v} = \frac{\partial \pi^*(w)}{\partial P(v)}$. Since $\pi^*$ is the stationary distribution of $M_P$, it satisfies the equilibrium condition $\pi^* = \pi^* M_P$. Differentiating,
\[
    \frac{\partial \pi^*}{\partial P(v)} = \frac{\partial \pi^*}{\partial P(v)} M_P + \pi^* \frac{\partial M_P}{\partial P(v)}.
\]
\end{proof}

Let $c_v^T = \frac{\partial \pi^*}{\partial P(v)}$ denote the row vector corresponding to the $v$th column of the Jacobian $J$, and let $r_v^T = \pi^* \frac{\partial M_P}{\partial P(v)}$ represent the direct perturbation vector. Rearranging the terms gives:
\begin{equation}
\label{eq:cvrv}
    c_v^T (I - M_P) = r_v^T
\end{equation}
To isolate $c_v^T$, we define 
$F = (I - M_P + \mathbf{1}\pi^*)^{-1}$,
which satisfies 
$(I - M_P) F = I - \mathbf{1}{\pi^*}$.  Multiplying both sides 
of \eqref{eq:cvrv} by $F$ on the right yields:
\[
    c_v^T (I - M_P) F = r_v^T F.
\]
Furthermore, since $c_v^T$ is the derivative of a probability distribution, its elements sum to zero, i.e., $c_v^T \mathbf{1} = 0$. This cancels the rank-one term $\mathbf{1}\pi^*$, yielding a  closed-form solution for the Jacobian column:
\[
    c_v^T = r_v^T F.
\]

%
We next show a bound on the norm of the Jacobian, in terms of the temperature.
\begin{lemma}
\label{lem:critical0}
There exist constants $C,\tau_0<\infty$ such that, for all $\tau\geq\tau_0$,
\[
    \sup_{P\in\Delta^{V-1}}\|J(P)\|\leq \frac{C}{\tau}.
\]
\end{lemma}
\begin{proof}
By \eqref{eq:attnoutputpd}, $\partial o_u/\partial P(v)$ is uniformly bounded over
$P\in\Delta^{V-1}$.  Since all remaining factors in \eqref{eq:partialM} except $1/\tau$ are bounded on the simplex, we obtain
\[
    \left\|\frac{\partial M_P}{\partial P(v)}\right\| = O\left( \frac{1}{\tau} \right).
\]

As $\tau\to\infty$, every row of $M_P$ converges uniformly in $P$ to the
uniform distribution $u$. Since $\pi^*(P)=\pi^*(P)M_P$,
\[
\|\pi^*(P)-u\|
\le \max_x \|M_P(x,\cdot)-u\|,
\]
so $\pi^*(P)\to u$ uniformly as well.
Hence $I-M_P+\mathbf{1}\pi^* \longrightarrow I$ uniformly in $P$, so
$
    F=(I-M_P+\mathbf{1}\pi^*)^{-1}
$
remains uniformly bounded for all sufficiently large $\tau$. 
Using $c_v^T=\pi^*(\partial M_P/\partial P(v))F$, and since
$\|\pi^*\|$ and $\|F\|$ are bounded, we obtain $\|c_v\| = O(1/\tau)$ for every column $c_v$ of $J$. Since there are only $V$ columns,
$\sup_P\|J(P)\|=O(1/\tau)$.
\end{proof}
Using these, we show that a 
temperature threshold exists, which is the main result in this section.
\begin{theorem}
\label{thm:critical}
There is a temperature
$\tau_c$ such that $P\mapsto\pi^*(P)$ is a
contraction for all $\tau>\tau_c$.  Hence, it has a unique fixed point
$P^*$, which is globally attracting under the ODE.
\end{theorem}
\begin{proof}
Choose $\tau_c > \max\{C, \tau_0\}$, where $C, \tau_0$ are
the constants in Lemma~\ref{lem:critical0}. Since the
simplex is convex, $P\mapsto\pi^*(P)$ is  a contraction and has
a unique fixed point $P^*$. Writing the ODE as
$\frac{dP}{dt}=\pi^*(P)-P$, 
 let $e(t)=P(t)-P^*$. Since $\pi^*(P^*)=P^*$ and
using
\[
\|\pi^*(P)-\pi^*(P^*)\|
\le q\|P-P^*\|,
\enspace 
q=\sup_P\|J(P)\|<1,
\]
we have
\smallmath{
\begin{equation*}
\frac{1}{2}\frac{d}{dt}\|e(t)\|^2
=
\langle e(t),\pi^*(P(t))-\pi^*(P^*)-e(t)\rangle
\le -(1-q)\|e(t)\|^2.
\end{equation*}
}
Integrating this inequality,
\[
\|P(t)-P^*\|\le e^{-(1-q)t}\|P(0)-P^*\|,
\]
so $P^*$ is globally attracting.  Since $\|P(t)-P^*\|$ decreases
strictly away from $P^*$, $\{P^*\}$ is the only ICT set of the ODE.
\end{proof}

%
Combining Lemma~\ref{lem:odeict} and Theorem~\ref{thm:critical}, for every $\tau>\tau_c$, we have $\lim_{t \rightarrow \infty} P_t = P^*$ (a.s.).  Moreover, as $\tau\to\infty$, $P^*$ converges to the uniform distribution,
since $M_P$ converges uniformly in $P$ to the transition matrix with every
row uniform.  For $\tau \leq \tau_c$, however, the above argument no
longer guarantees uniqueness of the limiting set.  Multiple limiting ICT sets are not ruled out and the behavior in this below-threshold temperature regime may depend on the specific model parameters.

\subsection{A Simple Example}
Let us consider a trivial example that shows three surprising facts about this simple transformer: 
\begin{enumerate}[nosep]
    \item It may non-determinstically approach one of multiple limits.
    \item The likelihood of each limit may depend on the starting conditions.
    \item As the temperature increases, the same transformer will transition into approaching a single limit, independent of starting conditions.
\end{enumerate} 
Our example has only two tokens, and the hidden dimension is one, so embeddings and weight matrices are simply scalars. For $a\in\mathcal{R}^+$, the model is fully specified by these values:
\[
E_{\rm in}[1]=1,\qquad E_{\rm in}[2]=-1,\qquad
W_Q=W_K=0,\qquad W_V=1,\qquad
E_{\rm out}[1]=a,\qquad E_{\rm out}[2]=-a.
\]
Let $p=P(1)$ be the fraction of token 1 in the history. Uniform attention gives
$o(P)=2p-1$, and every row of $M_P$ assigns probability
\[
q(p)=
\frac12\left(
1+\tanh\left(\frac{a}{\tau}(2p-1)\right)
\right)
\]
to token $1$.  Hence $\pi^*(P)=(q(p),1-q(p))$ and, with
$x=2p-1$, the limiting ODE becomes
\[
\frac{dx}{dt}
=
\tanh\left(\frac{a}{\tau}x\right)-x.
\]
For $\tau>a$, the right-hand map is a contraction and $x=0$ is
the unique globally attracting fixed point.  For $\tau<a$, $x=0$
is unstable and two symmetric attracting fixed points
$\pm x_*$ appear.  Thus even this two-token example exhibits the
temperature-dependent transition underlying Theorem~\ref{thm:critical}. Note that~\cite{mori2015correlation,hisakado2012two} present a similar analysis exhibiting the same variation in steady state.

Despite the interesting dynamics of the steady state, the historical token distribution for these models does approach a limit for every positive temperature, even if this limit might be a random variable. In Section~\ref{sec:nonconvergence}, we show that even this is not guaranteed by giving an explicit three-token construction with a proof of non-convergence of the token distribution.

\begin{figure}[t]
\centering
\begin{tikzpicture}


\begin{axis}[
    name=leftplot,
    width=0.47\textwidth,
    height=0.40\textwidth,
    xmin=-1.25,xmax=1.25,
    ymin=-1.25,ymax=1.25,
    axis lines=middle,
    xlabel={$x$},
    ylabel={$y$},
    title={High temperature: $\tau>a$},
    title style={font=\bfseries},
    xtick={-1,0,1},
    ytick={-1,0,1},
    legend style={
        at={(0.03,0.97)},
        anchor=north west,
        draw=black!30,
        fill=white,
        font=\small
    },
    clip=false
]

\addplot[
    domain=-1.2:1.2,
    samples=200,
    black!60,
    thick
]
{x};
\addlegendentry{$y=x$}

\addplot[
    domain=-1.2:1.2,
    samples=200,
    blue,
    very thick
]
{tanh(0.65*x)};
\addlegendentry{$y=\tanh((a/\tau)x)$}

\addplot[only marks,mark=*,mark size=2.2pt,black]
coordinates {(0,0)};

\node[anchor=west] at (axis cs:0.06,-0.08) {\small stable};

\draw[-{Latex[length=2mm]},thick]
    (axis cs:-0.95,-1.05) -- (axis cs:-0.25,-1.05);
\draw[-{Latex[length=2mm]},thick]
    (axis cs:0.95,-1.05) -- (axis cs:0.25,-1.05);

\node at (axis cs:0,-1.38)
    {\small unique globally attracting equilibrium};

\end{axis}

\begin{axis}[
    name=rightplot,
    at={(leftplot.outer east)},
    anchor=outer west,
    xshift=1.2cm,
    width=0.47\textwidth,
    height=0.40\textwidth,
    xmin=-1.25,xmax=1.25,
    ymin=-1.25,ymax=1.25,
    axis lines=middle,
    xlabel={$x$},
    ylabel={$y$},
    title={Low temperature: $\tau<a$},
    title style={font=\bfseries},
    xtick={-1,0,1},
    ytick={-1,0,1},
    legend style={
        at={(0.03,0.97)},
        anchor=north west,
        draw=black!30,
        fill=white,
        font=\small
    },
    clip=false
]

\addplot[
    domain=-1.2:1.2,
    samples=200,
    black!60,
    thick
]
{x};
\addlegendentry{$y=x$}

\addplot[
    domain=-1.2:1.2,
    samples=200,
    red,
    very thick
]
{tanh(1.5*x)};
\addlegendentry{$y=\tanh((a/\tau)x)$}

\def\xstar{0.85856}

\addplot[only marks,mark=*,mark size=2.2pt,black]
coordinates {(-\xstar,-\xstar) (\xstar,\xstar)};

\addplot[
    only marks,
    mark=o,
    mark size=2.5pt,
    black,
    fill=white
]
coordinates {(0,0)};

\draw[dashed,black!50]
    (axis cs:-\xstar,-\xstar) -- (axis cs:-\xstar,-1.10);
\draw[dashed,black!50]
    (axis cs:\xstar,\xstar) -- (axis cs:\xstar,-1.10);

\node[anchor=south east]
    at (axis cs:-\xstar,-\xstar) {\small stable};

\node[anchor=west]
    at (axis cs:\xstar,\xstar) {\small stable};

\node[anchor=west]
    at (axis cs:0.05,-0.10) {\small unstable};

\node at (axis cs:-\xstar,-1.18) {$-x_*$};
\node at (axis cs:0,-1.18) {$0$};
\node at (axis cs:\xstar,-1.18) {$x_*$};

\draw[-{Latex[length=2mm]},thick]
    (axis cs:-1.15,-1.05) -- (axis cs:-0.98,-1.05);

\draw[-{Latex[length=2mm]},thick]
    (axis cs:-0.25,-1.05) -- (axis cs:-0.70,-1.05);

\draw[-{Latex[length=2mm]},thick]
    (axis cs:0.25,-1.05) -- (axis cs:0.70,-1.05);

\draw[-{Latex[length=2mm]},thick]
    (axis cs:1.15,-1.05) -- (axis cs:0.98,-1.05);

\node at (axis cs:0,-1.38)
    {\small two stable equilibria and one unstable equilibrium};

\end{axis}

\end{tikzpicture}

\caption{
A two-token example illustrating the temperature transition:  $x = 2p-1$ and
$dx/dt = \tanh((a/\tau)x) - x$.  
For $\tau>a$, the uniform state is the unique globally attracting
equilibrium. For $\tau<a$, it becomes unstable and two symmetric
stable equilibria appear.
}
\label{fig:two-token-transition}
\end{figure}
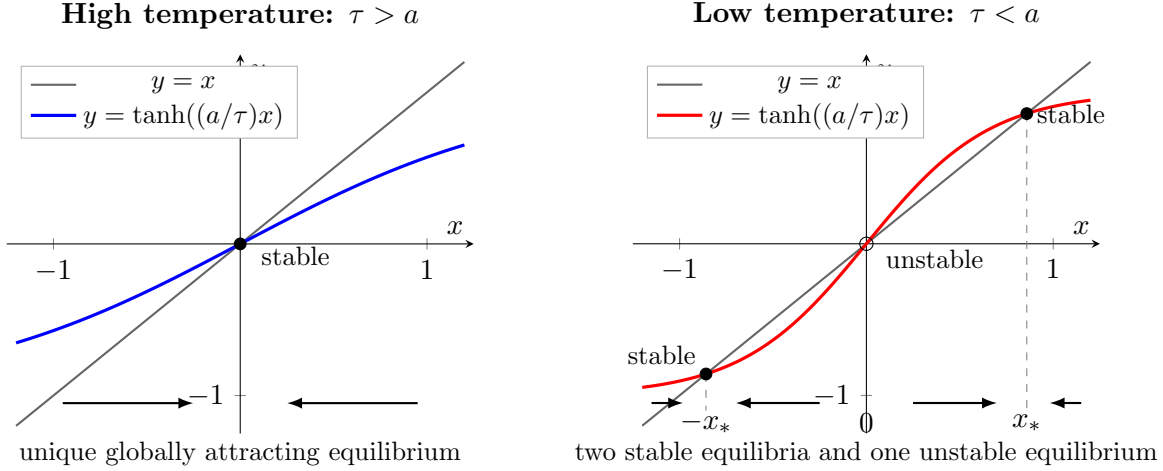




\section{Multi-Layer Transformer}
\label{sec:multi}

The analysis for a single-layer transformer
uses the fact that, without position encodings,
the history is summarized by the empirical token distribution. This does not
hold for a multi-layer transformer, where the key and value vectors stored at
a historical position depend on the \emph{context} in which that position was
generated. Nevertheless, for a fixed-width, fixed-depth transformer, each
historical position contributes only a fixed-dimensional vector to the global
KV cache. We therefore represent the unbounded history by an empirical
distribution over these vectors.

Throughout this section, the number of
layers and heads is fixed, and every local attention layer is causal with a
fixed window of width at most $w$.  Besides the global KV cache, the model
has no recurrent state other than the finite KV caches used to power the local layers.  As in a standard transformer block, the keys and values stored by an
attention layer are computed from the input to that layer, before its
attention output is produced. 
All projections,
MLPs, residual maps, and normalization maps are assumed to be uniformly Lipschitz on the reachable set and
all reachable hidden states, queries, keys, values, local cache states, and
pre-temperature output logits are uniformly bounded.  All constants below may depend on the 
architecture and its weights, but not on the generation time.  For local RoPE layers, attention depends only on relative offsets within
the fixed local window, so no absolute phase state is needed.

We consider a transformer with both \emph{global} attention layers,
whose attention window has the full history, and \emph{local} attention
layers with a fixed window. 
Let $L_{\rm glob}$ and $L_{\rm loc}$  be the numbers of global and local
attention layers respectively, each with $h$ heads of dimension $d$. 

Let
$\btheta_{\ell,a}=(\btheta_{\ell,a,1},\ldots,\btheta_{\ell,a,d/2})$ be the RoPE
frequencies of head $(\ell,a)$, and define the RoPE rotation as:
\[
    R(p\btheta_{\ell,a})
    :=
    \operatorname{diag}\bigl(
    R(p\btheta_{\ell,a,1}),\ldots,
    R(p\btheta_{\ell,a,d/2})\bigr),
\]
where
\[
    R(\alpha)=
    \begin{pmatrix}
    \cos\alpha&-\sin\alpha\\
    \sin\alpha&\cos\alpha
    \end{pmatrix}.
\]
For each historical position $p$, let $k^{\rm raw}_{\ell,a,p}$ be the
unrotated key, and let
\[
    k^{\rm rot}_{\ell,a,p}
    =R(p\btheta_{\ell,a}) \cdot k^{\rm raw}_{\ell,a,p},
\]
be the rotated key. The collection of global cache quantities generated at
position $p$ is
\[
    z_p=
    \bigl(k^{\rm rot}_{\ell,a,p},v_{\ell,a,p}\bigr)_{\ell\in[L_{\rm glob}],\,a\in[h]}
    \in\mathbb R^{2dhL_{\rm glob}}.
\]
Let $\mathcal Z$ be the closure of the set of all reachable $z_p$; by the boundedness assumption, $\mathcal Z$
is compact.


Let $\btheta$ denote the collection of RoPE frequencies across all global
heads, and let
$
    \omega_t=t\btheta\pmod{2\pi}
$
denote the vector of current RoPE phases. 
Let
\[
\eta_t=\frac1t\sum_{p=1}^t\delta_{z_p}
\]
be the empirical distribution of global cache vectors, where $\delta_z$ denotes a unit point mass at $z$.  Thus, the global cache changes slowly: each new token changes its empirical distribution by only $O(1/t)$. 

For each $z\in\mathcal Z$, let $k^{\rm rot}_{\ell,a}(z)$ and $v_{\ell,a}(z)$ denote its
key and value coordinates for global layer $\ell$ and head $a$. For a query
$
    q^{\rm rot}_{\ell,a,t}
    =R(t\btheta_{\ell,a}) \cdot q^{\rm raw}_{\ell,a,t},
$
the corresponding global attention output is
\smallmath{
\begin{equation}
\label{eq:attnintegral}
    o_{\ell,a}(t)
    =
    \frac{
        \displaystyle
        \int_{\mathcal Z}
        v_{\ell,a}(z) \cdot 
        \exp\!\left(
            \frac{\langle q^{\rm rot}_{\ell,a,t},k^{\rm rot}_{\ell,a}(z)\rangle}{\sqrt d}
        \right)d\eta_t(z)}
    {
        \displaystyle
        \int_{\mathcal Z}
        \exp\!\left(
            \frac{\langle q^{\rm rot}_{\ell,a,t},k^{\rm rot}_{\ell,a}(z)\rangle}{\sqrt d}
        \right)d\eta_t(z)
    }.
\end{equation}
}
Since the reachable queries and keys are bounded, the denominator in
\eqref{eq:attnintegral} is bounded away from zero uniformly in $t$.

Thus, once the RoPE rotation is incorporated into the stored key, its
historical location is no longer needed.
%
To characterize long-run RoPE phases, let
\[
    H_{\btheta}
    :=
    \overline{\{n\btheta\pmod{2\pi}:n\in\mathbb Z\}},
\]
be the closure of phases visited in the limit. Let $\mu_\btheta$ denote the
uniform probability distribution on $H_\btheta$.  We now show that averages along the RoPE phases converge to the average under
$\mu_\btheta$.
%
\begin{lemma}
\label{lem:tointegral}
For any continuous function $f:H_{\btheta}\to\mathbb C$,
\[
\sup_{\alpha\in H_\btheta}
\left|
\frac1T\sum_{n=1}^T f(\alpha+n\btheta)
-
\int_{H_\btheta}f(\omega)\,d\mu_\btheta(\omega)
\right|
\longrightarrow0.
\]
\end{lemma}
\begin{proof}
Consider $\alpha = 0$.  For an integer vector $m$ of the same dimension as $\btheta$, let $\chi_m(\omega)=e^{i\langle m,\omega\rangle}$.  Then, 
\[
\frac1T\sum_{n=1}^T \chi_m(n\btheta)
\longrightarrow
\begin{cases}
1, & \langle m,\btheta\rangle\in2\pi\mathbb Z,\\
0, & \text{otherwise}.
\end{cases}
\]
The average of the same function under the uniform distribution on
$H_\btheta$ has the same value. Since finite linear combinations of
$\chi_m$ uniformly approximate every continuous function on
$H_\btheta$, the result follows.  The same calculation with 
$\alpha+n\btheta$ instead of $n\btheta$
differs only by the factor $e^{i\langle m,\alpha\rangle}$, whose
absolute value is one, so the convergence is uniform in $\alpha$.
\end{proof}
To understand the impact of the local layers, note that each such layer has width at most $w$, and so its output can introduce dependence on at
most $w-1$ additional earlier token positions.
Thus, once the global attention outputs are fixed, the remaining computation
depends only on the most recent
$W=1+L_{\rm loc}(w-1)$ tokens.
  Let 
$u_t=(x_{t-W+1},\ldots,x_t)\in\Sigma^W$
be this $W$-gram (with fixed padding
 when $t<W$). Any fixed transformer in our model may be viewed as a function mapping from $u_t$ and the global cache state to a new $u_{t+1}$ and global KV cache column output.

\subsection{Frozen Cache Dynamics}
\label{sec:frozencache}

In this section, we first hold the global cache fixed
and study the evolution of $u_t$ over a block of $k$ steps. The motivation for this approach is as follows. Each new output modifies the global cache by only $O(1/t)$, 
so for $k = o(t)$, $\sup_{0\leq j\leq k} d_1(\eta_{t+j},\eta_t)=O(k/t) = o(1)$; thus the cache does indeed behave as if frozen. Once we understand the frozen cache behavior, we study how the drift during the frozen regime will eventually result in a new global cache state (possible, as $\sum_{t'>t}1/t'$ diverges), and what the trajectory over these evolving global cache states will look like.

Let $\eta$ be a probability distribution on
$\mathcal Z$ representing the (frozen) global cache state.  
With the global cache fixed, the recent $W$ tokens and the current RoPE
phase determine the next step.
For $u\in\Sigma^W$ and phase $\omega\in H_\btheta$, let $q_{\eta,\omega}(a\mid u)= \Pr [x_{t+1}=a\mid u_t=u,\omega_t=\omega,\eta]$
be the next token distribution produced by the transformer.   For $u,u'\in\Sigma^W$, define the transition matrix:
\[
    Q_{\eta,\omega}(u,u')
    =
    \Pr [u_{t+1}= u' \mid u_t=u,\omega_t=\omega,\eta].
\]
Thus $Q_{\eta,\omega}(u,u') > 0$ only if $u'$ is obtained from $u$ by dropping its first symbol and appending one new token. We first establish a uniform lower bound on next token probabilities.
This will imply that every $W$-gram can be reached from every other
in $W$ steps and hence the product
of the transition matrices contract.
    
We assume that $\tau\geq \tau_{\min}>0$.  
By the boundedness assumption, there is a constant $B<\infty$ such that
every pre-temperature output logit satisfies $|\ell^{\rm out}_a(\eta,u,\omega)|\leq B$
for all $u\in\Sigma^W$, $\omega\in H_\btheta$, and $a\in\Sigma$.
Therefore,
\begin{align}
q_{\eta,\omega}(a\mid u)
& \geq
\epsilon
:=
\frac{\exp(-2B/\tau_{\min})}{V}
>0.
\label{eq:eps}
\end{align}
Since the phase changes by $\btheta$ after each step, the
$m$-step transition matrix starting at phase $\omega$ is
\[
Q_{\eta,\omega}^{[m]}
:=
Q_{\eta,\omega}
\cdot 
Q_{\eta,\omega+\btheta}
\cdot 
\cdots
\cdot
Q_{\eta,\omega+(m-1)\btheta}.
\]

Starting from any $W$-gram, any particular $W$-gram
$u'=(y_1,\ldots,y_W)$ can be reached in $W$ steps by generating
$y_1,\ldots,y_W$. By \eqref{eq:eps}, each of these tokens has conditional
probability at least $\epsilon$. Hence
\[
Q_{\eta,\omega}^{[W]}(u,u')
\geq
\epsilon^W,
\]
for every $u,u'\in\Sigma^W$, uniformly in $\eta$ and $\omega$.

Let $\beta= V^W \epsilon^W \in (0, 1]$. Since every entry of the
$W$-step transition matrix is at least $\epsilon^W$, for any two
probability distributions $\lambda,\lambda'$ on $\Sigma^W$, we have 
\begin{align}
\left\|
\lambda \cdot Q_{\eta,\omega}^{[m]}
-
\lambda' \cdot Q_{\eta,\omega}^{[m]}
\right\|_{\rm tv}
\leq
(1-\beta)^{\lfloor \frac{m}{W} \rfloor}
\|\lambda-\lambda'\|_{\rm tv}.
\label{eq:tv}
\end{align}

Thus, with the cache fixed, 
the effect of the initial $W$-gram decays geometrically.  Since the RoPE phase changes at each step, the process is governed
by a sequence of transition matrices rather than by one fixed matrix; we study its limiting behavior.
\begin{lemma}
\label{lem:piQ}
For every fixed $\eta$, (i) there is a unique family
$\{ \pi_\eta(\omega) \}_{\omega \in H_\btheta}$ of
probability distributions such that
$\pi_\eta(\omega+\btheta)
=
\pi_\eta(\omega) \cdot Q_{\eta,\omega}$
and (ii) for every initial distribution $\lambda$,
\[
\left\|
\lambda \cdot Q_{\eta,\omega}^{[m]}
-
\pi_\eta(\omega+m\btheta)
\right\|_{\rm tv}
\leq
(1-\beta)^{\lfloor m/W\rfloor}.
\]
\end{lemma}
\begin{proof}
Fix $\omega$ and any distribution $\lambda$ on
$\Sigma^W$. Consider
\begin{align}
\lambda \cdot 
Q_{\eta,\omega-m\btheta} \cdot
Q_{\eta,\omega-(m-1)\btheta} \cdot
\cdots \cdot
Q_{\eta,\omega-\btheta}.
\label{eq:chain}
\end{align}
If a longer product is used, its last $m$ transition matrices are the
same as those in \eqref{eq:chain}, but with a possibly different distribution at
their starting point. By \eqref{eq:tv}, changing that starting distribution
changes the result by at most
$(1-\beta)^{\lfloor m/W\rfloor}$. Hence \eqref{eq:chain} converges as
$m\to\infty$, and its limit does not depend on $\lambda$. Define this
limit to be $\pi_\eta(\omega)$.

Shifting $\omega$ by $\btheta$ in \eqref{eq:chain} gives (i). Finally, apply
\eqref{eq:tv} to $\lambda$ and $\pi_\eta(\omega)$; (ii) now follows from (i). 
Uniqueness also follows from the same bound.
\end{proof}
Since only a fixed number of recent positions enter the local computation
and all computations are Lipschitz, replacing the cache states used at those
positions by the current cache $\eta_t$ changes the next step distribution by
$O(1/t)$. Recalling $\mathcal F_t$ is the history up to time $t$, 
\[
\left\|
\Pr[u_{t+1}\in\cdot\mid\mathcal F_t]
-
Q_{\eta_t,\omega_t}(u_t,\cdot)
\right\|_{\rm tv}
=
O(1/t).
\]
We next show that this frozen cache description remains accurate
over time intervals on which the empirical cache changes only slightly.
\begin{lemma}
\label{lem:frozen}
(i) There is a constant $L_\pi$ such that
\[
\sup_{\omega\in H_\btheta}
\|\pi_\eta(\omega)-\pi_{\eta'}(\omega)\|_{\rm tv}
\le
L_\pi \cdot d_1(\eta,\eta').
\]
(ii) Moreover, let $k=k(p)$ satisfy $k\to\infty$ and $k^2/p \to 0$.
Then 
$\|
\Pr[u_{p+k}\in\cdot\mid\mathcal F_p]
-
\pi_{\eta_p}(\omega_{p+k})
\|_{\rm tv}
\longrightarrow0.$
\end{lemma}
\begin{proof}
We first prove (i). 
Under our assumptions, there is a
constant $L_Q$ such that
\begin{align}
\sup_{u,\omega}
\left\|
Q_{\eta,\omega}(u,\cdot)
-
Q_{\eta',\omega}(u,\cdot)
\right\|_{\rm tv}
& \leq
L_Q \cdot d_1(\eta,\eta').
\label{eq:Qsmooth}
\end{align}
Indeed, for the first global layer this follows directly from \eqref{eq:attnintegral}:
its terms are uniformly bounded and Lipschitz in $z$, and the
denominator is uniformly bounded away from zero. The resulting attention
output therefore changes by $O(d_1(\eta,\eta'))$. Repeating this argument
through the fixed number of layers, using the Lipschitz assumptions,
proves \eqref{eq:Qsmooth}.

Combining \eqref{eq:Qsmooth} with \eqref{eq:tv} gives (i): comparing the products defining
$\pi_\eta(\omega)$ and $\pi_{\eta'}(\omega)$ term by term, a change
$r$ steps in the past is reduced geometrically in $r$, and summing
these changes gives the stated bound. Similarly, continuity of
$Q_{\eta,\omega}$ in $\omega$ implies continuity of
$\pi_\eta(\omega)$ in $\omega$.

We now return to the actual cache, which changes after every step $t$.  From the definition of $\eta_t$,
\[
\eta_{t+1}
=
\eta_t+\frac1{t+1}
\bigl(\delta_{z_{t+1}}-\eta_t\bigr).
\]
Since each cache update has size $O(1/t)$,
\begin{align*}
\sup_{0\leq j\leq k}
d_1(\eta_{p+j},\eta_p)
& =
O\!\left(\frac{k}{p}\right).
\end{align*}
Therefore, by \eqref{eq:Qsmooth},
\[
\sup_{0\leq j<k}\sup_u
\|Q_{\eta_{p+j},\omega_{p+j}}(u,\cdot)
 -
Q_{\eta_p,\omega_{p+j}}(u,\cdot)\|_{\rm tv}
=
O\!\left(\frac{k}{p}\right).
\]
Comparing the two $k$-step products one step at a time, the errors
from replacing $\eta_{p+j}$ by $\eta_p$ add up to $O(k^2/p)$.  The $O(1/t)$ error in the actual one-step transition
accumulates to $O(k/p)$, which is smaller. Hence, conditioned on
$\mathcal F_p$, the distribution of $u_{p+k}$ differs by
$O(k^2/p)$ from the distribution obtained by holding the cache fixed
at $\eta_p$. Lemma~\ref{lem:piQ} thus gives 
\smallmath{
\begin{align*}
\left\|
\Pr[u_{p+k}\in\cdot\mid\mathcal F_p]
-
\pi_{\eta_p}(\omega_{p+k})
\right\|_{\rm tv}
\leq
(1-\beta)^{\lfloor \frac{k}{W} \rfloor}
+
O\!\left(\frac{k^2}{p}\right).
\end{align*}
}
Both terms vanish under $k\to\infty$ and $k^2/p \to 0$, proving (ii).
\end{proof}
Thus, over a block that is long enough for the recent tokens to forget
their initial state but short enough that the global cache changes little,
their distribution is determined asymptotically by the cache state at the
start of the block and the current RoPE phase.   Although the instantaneous distribution may depend on the RoPE phase,
Lemma~\ref{lem:tointegral} allows this phase dependence to be averaged over a long block.

\subsection{Slow Cache Dynamics}
\label{sec:slowcache}

Lemma~\ref{lem:frozen} describes the recent token distribution while the global cache
changes only slightly. We now use this to study the
average evolution of the cache itself.

For a fixed $\eta$, $u\in\Sigma^W$, $\omega\in H_\btheta$, and a
generated token $a\in\Sigma$, let $G(\eta,u,\omega,a)\in \mathcal{Z}$
denote the new global cache vector inserted at the next position,
whose RoPE phase is $\omega+\btheta$. The 
Lipschitz assumptions above mean that 
$G$ is Lipschitz in $\eta$ with respect to $d_1$ and in $\omega$.
Define 
\smallmath{
\[
\Gamma(\eta)
:=
\int_{H_\btheta}
\sum_{\stackrel{u\in\Sigma^W}{a\in\Sigma}}
\pi_\eta(\omega)(u) \cdot
q_{\eta,\omega}(a\mid u) \cdot
\,\delta_{G(\eta,u,\omega,a)}
\,d\mu_\btheta(\omega).
\]
}
Thus $\Gamma(\eta)$ is the average distribution of the next cache vector
when the recent tokens have adjusted to the frozen cache $\eta$ and the
RoPE phase is averaged.

Since the empirical cache updates with step size $1/(t+1)$, its natural 
continuous time scale is
$s_t:=\sum_{j=1}^t\frac1j = \log t + O(1)$; we
use $s$ for this slow time variable.  
The corresponding equation for the
empirical cache is 
\begin{align}
\frac{d\eta}{ds}
& =
\Gamma(\eta)-\eta.
\label{eq:slowcache}
\end{align}
Since $\eta_s$ is a probability distribution rather than a finite-dimensional vector, we interpret \eqref{eq:slowcache} by 
testing it against Lipschitz functions: for every $\varphi: \mathcal{Z} \to\mathbb R$,
\[
\frac{d}{ds}\int_{\mathcal Z} \varphi(z)\,d\eta_s(z)
=
\int_{\mathcal Z} \varphi(z)\,d\Gamma(\eta_s)(z)
-
\int_{\mathcal Z} \varphi(z)\,d\eta_s(z).
\]
For a Lipschitz $\varphi: \mathcal{Z} \to\mathbb R$, let 
\begin{align*}
\eta(\varphi) & :=\int_{\mathcal Z} \varphi(z)\,d\eta(z), \\
G_\varphi(\eta,u,\omega)
& := \sum_{a\in\Sigma}
q_{\eta,\omega}(a\mid u) \cdot 
\varphi(G(\eta,u,\omega,a)).
\end{align*}
Thus, $\eta(\varphi)$ is the average
value of $\varphi$ under the current cache
distribution and $G_\varphi(\eta, u, \omega)$ is the expected value of $\varphi$ on the
next cache entry, given frozen cache, current local state, and current phase.  The next theorem shows that, over a long block on which the cache changes only slightly, the average change of $\eta(\varphi)$ is
asymptotically
$\Gamma(\eta)(\varphi)-\eta(\varphi)$.
\begin{theorem}
\label{thm:multi}
Let $k=k(p)$ satisfy 
$k\to\infty$ and $k^2/p \to 0.$  
Then, for every Lipschitz $\varphi: {\mathcal Z} \to\mathbb R$,
\smallmath{
\[
\frac1k\sum_{j=0}^{k-1}
\mathbb E\!\left[
\varphi(z_{p+j+1})-\eta_{p+j}(\varphi)
\,\middle|\,\mathcal F_p
\right]
=
\Gamma(\eta_p)(\varphi)-\eta_p(\varphi)+o(1),
\]
}
where
\[
\Gamma(\eta)(\varphi)
:=
\int_{\mathcal Z} \varphi(z)\,d\Gamma(\eta)(z).
\]
\end{theorem}
\begin{proof}
As noted before Lemma~\ref{lem:frozen},
\[
\mathbb E[
\varphi(z_{t+1})\mid\mathcal F_t]
=
G_\varphi(\eta_t,u_t,\omega_t)+O(1/t).
\]
Over a block of $k$ steps beginning at time $p$,
\[
d_1(\eta_{p+j},\eta_p)=O(k/p),
\qquad 0\leq j\leq k.
\]
By the Lipschitz bounds above, we may therefore replace
$\eta_{p+j}$ by $\eta_p$ at a total averaged cost $O(k/p)$.

The same comparison used in the 
proof of Lemma~\ref{lem:frozen} gives, uniformly for $0\leq j\leq k$,
\smallmath{
\[
\left\|
\Pr[u_{p+j}\in\cdot\mid\mathcal F_p]
-
\pi_{\eta_p}(\omega_{p+j})
\right\|_{\rm tv}
\leq
C\left(
(1-\beta)^{\lfloor j/W\rfloor}
+\frac{j^2}{p}
\right),
\]
}
for a constant $C$. Averaging over $j$ therefore gives
\smallmath{
\begin{align}
& \frac1k\sum_{j=0}^{k-1}
\mathbb E[
G_\varphi(\eta_p,u_{p+j},\omega_{p+j})
\mid\mathcal F_p] \nonumber \\
& =
\frac1k\sum_{j=0}^{k-1}
\sum_{u\in\Sigma^W}
\pi_{\eta_p}(\omega_{p+j})(u)
G_\varphi(\eta_p,u,\omega_{p+j})
+o(1).
\label{eq:Gavg}
\end{align}
}

The function
\[
\omega\mapsto
\sum_{u\in\Sigma^W}
\pi_{\eta_p}(\omega)(u)
G_\varphi(\eta_p,u,\omega)
\]
is continuous.  These functions depend continuously on $\eta$ and $\omega$.  Since
the possible cache distributions form a compact set, this family can
be uniformly approximated by finitely many such functions.  Lemma~\ref{lem:tointegral} 
therefore applies
uniformly over this family and 
the RHS of \eqref{eq:Gavg} converges to
\[
\int_{H_\btheta}
\sum_{u\in\Sigma^W}
\pi_{\eta_p}(\omega)(u)
G_\varphi(\eta_p,u,\omega)
\,d\mu_\btheta(\omega)
=
\Gamma(\eta_p)(\varphi).
\]
Also, since
$d_1(\eta_{p+j},\eta_p)=O(k/p)$ for $0\le j\le k$,
\[
\frac1k\sum_{j=0}^{k-1}
\mathbb E[\eta_{p+j}(\varphi)\mid\mathcal F_p]
=
\eta_p(\varphi)+o(1).
\]
Combining these estimates completes the proof.
\end{proof}
Thus $\Gamma(\eta)-\eta$ gives the average change of the empirical
cache on the $s$-time scale; but
nothing here requires solutions of
\eqref{eq:slowcache} to converge.  Note that 
the bounds above also imply that $\Gamma$ is Lipschitz with respect to
$d_1$: for some $L_\Gamma<\infty$,
$
d_1(\Gamma(\eta),\Gamma(\eta'))
\leq
L_\Gamma \cdot d_1(\eta,\eta').
$
Indeed, this follows by comparing the corresponding factors
$\pi_\eta$, $q_{\eta,\omega}$, $G$ in the definition of $\Gamma$.  Thus \eqref{eq:slowcache} has a unique solution for any initial cache
distribution.

Let $\widehat\eta(s)$ interpolate the values
$\widehat\eta(s_t)=\eta_t$ linearly between consecutive $s_t$'s.  Theorem~\ref{thm:multi} identifies the average change of the cache over short blocks.
The next result shows that these local approximations combine to describe
the full cache trajectory on the $s$-time scale.

\begin{corollary}
\label{cor:slowcache}
Let $\eta^{(t)}(s)$ be the solution of
\eqref{eq:slowcache} with 
$\eta^{(t)}(0)=\eta_t$.  
Then, for every fixed $T<\infty$,
\[
\sup_{0\leq s\leq T}
d_1\!\left(
\widehat\eta(s_t+s),
\eta^{(t)}(s)
\right)
\longrightarrow0
\qquad\text{a.s.}
\]
Thus $\eta_t$ tracks \eqref{eq:slowcache} a.s.
\end{corollary}
\begin{proof}
For a Lipschitz function $\varphi: \mathcal{Z}\to\mathbb R$, the empirical cache
update gives
\[
\eta_{t+1}(\varphi)-\eta_t(\varphi)
=
\frac1{t+1}
\bigl(\varphi(z_{t+1})-\eta_t(\varphi)\bigr).
\]
By Theorem~\ref{thm:multi}, over blocks that grow to infinity but remain short
relative to the current time, the average conditional change is $\Gamma(\eta)(\varphi)-\eta(\varphi)+o(1),$
where $\eta$ is the cache state at the start of the block.  Since the
cache changes by $o(1)$ across such a block and $\Gamma$ is Lipschitz,
we can use the current cache state within the block.  Partitioning any fixed interval of the $s$-time scale into such blocks,
Theorem~\ref{thm:multi} therefore shows that the accumulated difference between the
conditional change and $\Gamma(\eta)-\eta$ tends to zero.

The remaining random term is bounded and has conditional mean zero.
After multiplying by $1/(t+1)$, its weighted sum converges (a.s.) because $\sum_t\frac{1}{(t+1)^2}<\infty$.
The $O(1/t)$ approximation error used in Section~\ref{sec:frozencache} contributes only
$O(1/t^2)$ after multiplication by the step size, and is therefore
summable as well.

Since $\mathcal{Z}$ is compact, controlling finitely many $1$-Lipschitz test
functions up to an arbitrarily small error is enough to control $d_1$; hence the same bounds hold uniformly in the definition of $d_1$.
Consequently, over every fixed interval of the $s$-time scale,
$\widehat\eta$ satisfies
\[
\widehat\eta(s_t+s)
=
\eta_t
+
\int_0^s
\bigl(
\Gamma(\widehat\eta(s_t+r))
-
\widehat\eta(s_t+r)
\bigr)\,dr
+
o(1),
\]
where the error tends to zero uniformly for $0\leq s\leq T$ in the
distance $d_1$.

The exact solution $\eta^{(t)}$ satisfies the same equation without the
$o(1)$ term. Since $\Gamma$ is Lipschitz, their distance is at most
\[
o(1)
+
(L_\Gamma+1)
\int_0^s
d_1\!\left(
\widehat\eta(s_t+r),
\eta^{(t)}(r)
\right)\,dr.
\]
Repeated substitution in this inequality shows that, uniformly for
$0\leq s\leq T$,
\[
d_1\!\left(
\widehat\eta(s_t+s),
\eta^{(t)}(s)
\right)
\leq
o(1)e^{(L_\Gamma+1)T};
\]
the RHS vanishes (a.s).
\end{proof}
%

\subsection{High-Temperature Convergence}
\label{sec:high-temp-multilayer}

Corollary~\ref{cor:slowcache} shows that the empirical cache
tracks \eqref{eq:slowcache}, but does not imply that solutions of \eqref{eq:slowcache} converge.
We now show that convergence does hold at sufficiently high
temperature for every fixed transformer.

In this section we make the dependence on the  temperature $\tau$ 
explicit, writing
$q^{(\tau)}_{\eta,\omega}$, $Q^{(\tau)}_{\eta,\omega}$,
$\pi^{(\tau)}_\eta$, and $\Gamma_\tau$.
Let $z=(z^{(1)},\ldots,z^{(L_{\rm glob})})$
be the cache vector where $z^{(j)}$ contains the keys and values stored by the $j$th
global layer, and let $\eta^{(j)}$ be the corresponding marginal of
$\eta$.  Likewise, let 
$G(\eta,u,\omega,a)
=
\bigl(G^{(1)}(\eta,u,\omega,a),\ldots,
G^{(L_{\rm glob})}(\eta,u,\omega,a)\bigr).$

For $0<\delta<1$, define
\[
\|z-z'\|_\delta
:=
\sum_{j=1}^{L_{\rm glob}}
\delta^{j-1}\|z^{(j)}-z'^{(j)}\|,
\]
and let $d_{1,\delta}$ denote the $1$-Wasserstein distance induced by
this norm. Since the number of layers is finite, $d_{1,\delta}$ and
$d_1$ are equivalent.
We can now show the following contraction result on the expected next cache state for two different histories:
\begin{lemma}\label{lem:gamma-contraction}
For every small $\delta>0$, there are $A(\delta)$ and
$C_\delta$ with $A(\delta) \to 0$ as $\delta \to 0$ such that
\begin{align}
d_{1,\delta}\bigl(\Gamma_\tau(\eta),\Gamma_\tau(\eta')\bigr)
& \leq
\left(
A(\delta)+\frac{C_\delta}{\tau}
\right)
d_{1,\delta}(\eta,\eta').
\label{eq:gamma-contraction}
\end{align}
\end{lemma}
\begin{proof}
By the ordering of the layers, $G^{(j)}$ can depend on the global cache
only through global layers $1,\ldots,j-1$.  Hence there are constants
$C_{ji}<\infty$ such that
\smallmath{
\[
\left\|
G^{(j)}(\eta,u,\omega,a)
-
G^{(j)}(\eta',u,\omega,a)
\right\|
\leq
\sum_{i<j}
C_{ji}\,
d_1^{(i)}(\eta^{(i)},\eta'^{(i)}),
\]
}
uniformly in $u,\omega,a$, where $d_1^{(i)}$ denotes the distance
$d_1$ on the cache coordinates of layer $i$.
This follows from
\eqref{eq:attnintegral} and the Lipschitz assumptions, applied  through the
layers; in particular, $G^{(1)}$ does not depend on $\eta$.

A function of $z^{(i)}$ that is $1$-Lipschitz in that coordinate is
$\delta^{-(i-1)}$-Lipschitz under $\|\cdot\|_\delta$.  Therefore,
\[
d_1^{(i)}(\eta^{(i)},\eta'^{(i)})
\leq
\delta^{-(i-1)} \cdot d_{1,\delta}(\eta,\eta').
\]
It follows that
\[
\left\|
G(\eta,u,\omega,a)
-
G(\eta',u,\omega,a)
\right\|_\delta
\leq
A(\delta) \cdot d_{1,\delta}(\eta,\eta'),
\]
where
\[
A(\delta)
:=
\sum_{j=2}^{L_{\rm glob}}
\sum_{i<j}
C_{ji} \cdot \delta^{\,j-i}.
\]
Thus $A(\delta)\to0$ as $\delta\to0$.

We next consider the next token distribution.  The argument used for
\eqref{eq:Qsmooth}, stopped just before the final softmax, shows that the
pre-temperature output logits are Lipschitz in $\eta$.  Since the
derivative of $\operatorname{softmax}(x/\tau)$ 
is $O(1/\tau)$, for every fixed $\delta$ there is a constant
$C_\delta<\infty$ such that, for all sufficiently large $\tau$,
\[
\sup_{u,\omega}
\left\|
q^{(\tau)}_{\eta,\omega}(\cdot\mid u)
-
q^{(\tau)}_{\eta',\omega}(\cdot\mid u)
\right\|_{\rm tv}
\leq
\frac{C_\delta}{\tau} \cdot 
d_{1,\delta}(\eta,\eta').
\]
The same bound therefore holds for the corresponding transition matrices $Q^{(\tau)}_{\eta,\omega}$.

For sufficiently large $\tau$, the lower bound in \eqref{eq:eps}, and hence the
contraction in \eqref{eq:tv}, can be chosen uniformly in $\tau$.  Repeating the
proof of Lemma~\ref{lem:piQ} therefore gives
\[
\sup_{\omega\in H_\btheta}
\left\|
\pi^{(\tau)}_\eta(\omega)
-
\pi^{(\tau)}_{\eta'}(\omega)
\right\|_{\rm tv}
\leq
\frac{C_\delta}{\tau} \cdot 
d_{1,\delta}(\eta,\eta'),
\]
after increasing $C_\delta$ if necessary.

We can now compare $\Gamma_\tau(\eta)$ and
$\Gamma_\tau(\eta')$.  

(i) If the weights 
$\pi^{(\tau)}_\eta(\omega)(u)
q^{(\tau)}_{\eta,\omega}(a\mid u)$
are kept fixed and only the point
$G(\eta,u,\omega,a)$ is replaced by
$G(\eta',u,\omega,a)$, the change is at most
$A(\delta) \cdot d_{1,\delta}(\eta,\eta').$

(ii) If the points are kept fixed and the weights are changed, the bounds
above give a change of at most
$\frac{C_\delta}{\tau} \cdot d_{1,\delta}(\eta,\eta')$.  Hence
\begin{align*}
d_{1,\delta}\bigl(\Gamma_\tau(\eta),\Gamma_\tau(\eta')\bigr)
& \leq
\left(
A(\delta)+\frac{C_\delta}{\tau}
\right)
d_{1,\delta}(\eta,\eta').
\qedhere
\end{align*}
\end{proof}
We have the ingredients to prove
high-temperature convergence.
\begin{theorem}
\label{thm:multi-high-temp}
For every transformer satisfying the assumptions of this
section, there is a finite $\tau_c$ such that, for every
$\tau>\tau_c$, $\Gamma_\tau$ has a unique fixed point
$\eta_\tau^*$.  Every solution of
$\frac{d\eta}{ds}=\Gamma_\tau(\eta)-\eta$
converges exponentially to $\eta_\tau^*$, and
$d_1(\eta_t,\eta_\tau^*)\longrightarrow0$
\text{a.s.}
\end{theorem}
\begin{proof}
Choose $\delta>0$ small enough that $A(\delta)<1/4$, and choose
$\tau_c$ large such that all the preceding large temperature bounds hold and
$\tau_c > 4 C_\delta$. 
By \eqref{eq:gamma-contraction}, for every $\tau>\tau_c$,
$\Gamma_\tau$ is a contraction under $d_{1,\delta}$; let $\kappa<1$ be its contraction
constant.  Since $\mathcal Z$ is compact, iterating
$\Gamma_\tau$ gives a unique fixed point $\eta_\tau^*$.

Let $\eta_s$ be any solution of the ODE.  Its integral form is
\[
\eta_s
=
e^{-s}\eta_0
+
\int_0^s e^{-(s-r)}\Gamma_\tau(\eta_r)\,dr.
\]
Using $\Gamma_\tau(\eta_\tau^*)=\eta_\tau^*$ and the contraction
bound,
\[
d_{1,\delta}(\eta_s,\eta_\tau^*)
\leq
e^{-s}d_{1,\delta}(\eta_0,\eta_\tau^*)
+
\kappa
\int_0^s
e^{-(s-r)}
d_{1,\delta}(\eta_r,\eta_\tau^*)\,dr,
\]
and repeated substitution gives
\begin{align}
d_{1,\delta}(\eta_s,\eta_\tau^*)
& \leq
e^{-(1-\kappa)s}
d_{1,\delta}(\eta_0,\eta_\tau^*).
\label{eq:ode-high-temp-contraction}
\end{align}
It remains to return to the stochastic cache process.  Since
$d_{1,\delta}$ and $d_1$ are equivalent, Corollary~\ref{cor:slowcache}
also gives uniform tracking in $d_{1,\delta}$ over every fixed interval
of the $s$-time scale.  Choose $T>0$ and
let $\rho:=e^{-(1-\kappa)T}<1$; choose
times $t_0<t_1<\cdots$ so that
$s_{t_{m+1}}-s_{t_m}\to T$.  Applying Corollary~\ref{cor:slowcache} on each 
interval and using \eqref{eq:ode-high-temp-contraction},
\[
d_{1,\delta}(\eta_{t_{m+1}},\eta_\tau^*)
\leq
\rho\,
d_{1,\delta}(\eta_{t_m},\eta_\tau^*)
+
o(1)
\qquad\text{a.s.}
\]
Hence
$d_{1,\delta}(\eta_{t_m},\eta_\tau^*)\to0$ (a.s.) The same bound on the intervals between consecutive $t_m$ yields 
$d_{1,\delta}(\eta_t,\eta_\tau^*)\longrightarrow0$
(a.s.)  Now, the equivalence of $d_{1,\delta}$ and $d_1$ completes the proof.
\end{proof}
Thus every fixed transformer has a sufficiently high-temperature
regime in which the empirical cache converges.  The threshold
$\tau_c$ may depend on the architecture and its weights.

\section{Non-Convergence}
\label{sec:nonconvergence}

We now give an example showing that even the simplified single-layer transformer of Section~\ref{sec:single} need not converge.  Let $\Sigma=\{1,2,3\}$, $d=2$, and temperature $\tau=1$.  Define
\[
v_1=(1,0),\qquad
v_2=\left(-\frac12,\frac{\sqrt3}{2}\right),\qquad
v_3=\left(-\frac12,-\frac{\sqrt3}{2}\right),
\]
and let $E_{\rm in}[i]=v_i$.  Set
\[
W_Q=W_K=0,\qquad W_V=I.
\]
All attention scores are then zero, so for every query token the
attention output is
\[
o(P)=\sum_{i=1}^3P(i)v_i.
\]

Fix a constant $c>6$ and choose the output embeddings cyclically as
\[
E_{\rm out}[1]=-\frac{2c}{3}v_2,\qquad
E_{\rm out}[2]=-\frac{2c}{3}v_3,\qquad
E_{\rm out}[3]=-\frac{2c}{3}v_1.
\]
Since
\[
\langle o(P),v_i\rangle=\frac32P(i)-\frac12,
\]
the output logits, up to a common additive constant, are
\[
-cP(2),\qquad -cP(3),\qquad -cP(1).
\]
Thus every row of $M_P$ is the same distribution $q(P)$, where,
with indices interpreted cyclically,
\[
q_i(P)
\propto
\exp(-cP(i+1)),
\qquad i=1,2,3.
\]
Consequently $\pi^*(P)=q(P)$ and the limiting ODE is
\[
\frac{dP}{dt}=q(P)-P.
\]
This ODE has a unique fixed point
\[
\bar P=\left(\frac13,\frac13,\frac13\right).
\]
Indeed, at any fixed point,
\[
P(i)>P(j)
\quad\Longleftrightarrow\quad
P(i+1)<P(j+1).
\]
If, for example, $P(1)>P(2)$, then $P(3)>P(2)$,
which implies $P(3)>P(1)$, which in turn implies
$P(2)>P(1)$, a contradiction.  Hence all three coordinates
must be equal.

The fixed point is unstable when $c>6$.  
The Jacobian of $q$ at $\bar P$, restricted to the tangent plane
$\{x:\mathbf 1^\top x=0\}$, is
\[
J_q(\bar P)=-\frac{c}{3}S,
\qquad
S(x_1,x_2,x_3)=(x_2,x_3,x_1).
\]
The two eigenvalues of the Jacobian of $q(P)-P$ on this plane are
\[
\frac{c}{6}-1
\ \pm\
i\,\frac{\sqrt3\,c}{6}.
\]
Their real parts are positive for $c>6$, so $\bar P$ is a
linearly unstable equilibrium.

Finally, the stochastic process itself cannot converge.  Since
$q_i(P)$ is uniformly bounded away from zero on the simplex, the
sampling noise is nondegenerate in every direction tangent to the
simplex.  The standard non-convergence result for linearly unstable
equilibria in stochastic approximation
\citep[Section~9]{benaim1999dynamics} therefore gives
\[
\Pr[P_t\to\bar P]=0.
\]
By Lemma~\ref{lem:odeict}, if $P_t$ converged, its limit would have to be a fixed
point of the ODE.  Since $\bar P$ is the only fixed point, $P_t$
does not converge almost surely.

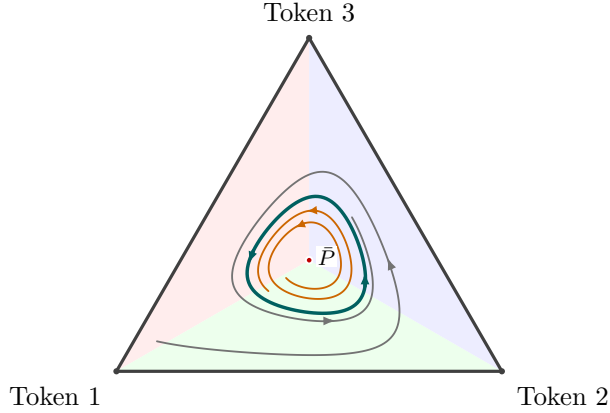
\begin{figure}[h]
\centering
\begin{tikzpicture}[scale=5.1,line cap=round,line join=round]
  \coordinate (P1) at (0,0);
  \coordinate (P2) at (1,0);
  \coordinate (P3) at (0.5,0.8660254);
  \coordinate (C) at (0.5,0.2886751);

  \fill[red!7]   (P1)--(P3)--(C)--cycle;   
  \fill[green!7] (P2)--(P1)--(C)--cycle;   
  \fill[blue!7]  (P3)--(P2)--(C)--cycle;   
  \draw[very thick,black!75] (P1)--(P2)--(P3)--cycle;

  \draw[orange!80!black,line width=0.7pt]
    plot[smooth] coordinates {
    (0.44000,0.24249) (0.44699,0.23592) (0.45527,0.23009) (0.46466,0.22507)
    (0.47495,0.22093) (0.48589,0.21772) (0.49722,0.21553) (0.50867,0.21439)
    (0.51998,0.21437) (0.53089,0.21552) (0.54117,0.21788) (0.55062,0.22149)
    (0.55906,0.22637) (0.56634,0.23252) (0.57235,0.23993) (0.57700,0.24853)
    (0.58025,0.25824) (0.58208,0.26892) (0.58251,0.28041) (0.58159,0.29247)
    (0.57939,0.30485) (0.57599,0.31728) (0.57150,0.32945) (0.56600,0.34108)
    (0.55961,0.35189) (0.55240,0.36161) (0.54446,0.37002) (0.53585,0.37689)
    (0.52663,0.38207) (0.51687,0.38543) (0.50664,0.38687) (0.49602,0.38635)
    (0.48510,0.38388) (0.47402,0.37953) (0.46292,0.37342) (0.45197,0.36573)
    (0.44138,0.35668) (0.43137,0.34652) (0.42215,0.33550) (0.41394,0.32389)
    (0.40695,0.31193) (0.40139,0.29984) (0.39743,0.28782) (0.39523,0.27602)
    (0.39491,0.26460) (0.39657,0.25367) (0.40028,0.24333) (0.40605,0.23366)
    (0.41384,0.22474) (0.42353,0.21664) (0.43494,0.20942) (0.44784,0.20314)
    (0.46190,0.19786) (0.47679,0.19365) (0.49215,0.19055) (0.50759,0.18864)
    (0.52278,0.18796) (0.53740,0.18860) (0.55118,0.19060) (0.56386,0.19403)
    (0.57525,0.19895) (0.58518,0.20540) (0.59351,0.21340) (0.60017,0.22297)
    (0.60508,0.23405) (0.60822,0.24655) (0.60963,0.26032) (0.60935,0.27514)
    (0.60749,0.29072) (0.60419,0.30674) (0.59957,0.32282) (0.59379,0.33861)
    (0.58699,0.35373) (0.57930,0.36785) (0.57080,0.38066) (0.56159,0.39190)
    (0.55172,0.40133) (0.54123,0.40875) (0.53015,0.41401) (0.51852,0.41699)
    (0.50636,0.41761) (0.49376,0.41584) (0.48078,0.41173) (0.46755,0.40539)
    (0.45424,0.39700) (0.44105,0.38681) (0.42822,0.37513) (0.41599,0.36228)
    (0.40462,0.34860) (0.39437,0.33441) (0.38547,0.31998) (0.37814,0.30557)
    (0.37257,0.29138) (0.36895,0.27756) (0.36742,0.26426) (0.36811,0.25156)
    (0.37112,0.23956) (0.37652,0.22831) (0.38430,0.21788) (0.39441,0.20830)
    };
  \draw[orange!80!black,line width=0.9pt,-{Latex[length=1.35mm,width=1.2mm]}] (0.48664,0.38387) -- (0.46292,0.37342);
  \draw[orange!80!black,line width=0.9pt,-{Latex[length=1.35mm,width=1.2mm]}] (0.515172,0.4180133) -- (0.49376,0.41584);

  \draw[black!55,line width=0.7pt]
    plot[smooth] coordinates {
    (0.10500,0.07794) (0.13045,0.07277) (0.15836,0.06799) (0.18866,0.06359)
    (0.22115,0.05956) (0.25541,0.05591) (0.29090,0.05264) (0.32701,0.04976)
    (0.36310,0.04729) (0.39863,0.04525) (0.43313,0.04368) (0.46626,0.04262)
    (0.49777,0.04211) (0.52749,0.04219) (0.55532,0.04294) (0.58119,0.04441)
    (0.60507,0.04667) (0.62696,0.04979) (0.64684,0.05387) (0.66473,0.05900)
    (0.68060,0.06528) (0.69447,0.07283) (0.70632,0.08179) (0.71612,0.09229)
    (0.72385,0.10448) (0.72946,0.11852) (0.73293,0.13452) (0.73424,0.15258)
    (0.73340,0.17273) (0.73048,0.19488) (0.72562,0.21881) (0.71900,0.24419)
    (0.71089,0.27055) (0.70157,0.29736) (0.69132,0.32408) (0.68041,0.35020)
    (0.66902,0.37529) (0.65733,0.39899) (0.64543,0.42103) (0.63337,0.44118)
    (0.62118,0.45927) (0.60885,0.47516) (0.59633,0.48873) (0.58358,0.49988)
    (0.57052,0.50850) (0.55708,0.51448) (0.54318,0.51773) (0.52873,0.51815)
    (0.51367,0.51566) (0.49797,0.51024) (0.48163,0.50191) (0.46471,0.49080)
    (0.44737,0.47715) (0.42981,0.46130) (0.41230,0.44369) (0.39516,0.42480)
    (0.37870,0.40511) (0.36321,0.38506) (0.34896,0.36501) (0.33618,0.34527)
    (0.32508,0.32605) (0.31581,0.30751) (0.30855,0.28977) (0.30341,0.27288)
    (0.30056,0.25689) (0.30011,0.24182) (0.30220,0.22768) (0.30694,0.21446)
    (0.31444,0.20217) (0.32475,0.19079) (0.33785,0.18033) (0.35364,0.17080)
    (0.37187,0.16219) (0.39221,0.15453) (0.41418,0.14784) (0.43726,0.14215)
    (0.46089,0.13749) (0.48454,0.13390) (0.50776,0.13144) (0.53015,0.13015)
    (0.55138,0.13009) (0.57122,0.13134) (0.58946,0.13397) (0.60596,0.13807)
    (0.62058,0.14374) (0.63324,0.15109) (0.64385,0.16019) (0.65235,0.17116)
    (0.65870,0.18404) (0.66287,0.19886) (0.66489,0.21556) (0.66483,0.23399)
    (0.66281,0.25392) (0.65900,0.27498) (0.65363,0.29675) (0.64693,0.31875)
    (0.63912,0.34051) (0.63041,0.36158) (0.62098,0.38155) (0.61094,0.40009)
    };
  \draw[black!55,line width=0.9pt,-{Latex[length=1.75mm,width=1.2mm]}] (0.713748,0.26788) -- (0.70157,0.29736);
  \draw[black!55,line width=0.9pt,-{Latex[length=1.75mm,width=1.2mm]}] (0.54454,0.13090) -- (0.57122,0.13134);

  \draw[teal!75!black,line width=1.3pt]
    plot[smooth cycle] coordinates {
    (0.50000,0.15274) (0.52282,0.15095) (0.54456,0.15079) (0.56481,0.15236)
    (0.58325,0.15577) (0.59961,0.16115) (0.61370,0.16861) (0.62535,0.17826)
    (0.63446,0.19021) (0.64096,0.20448) (0.64482,0.22103) (0.64611,0.23969)
    (0.64495,0.26015) (0.64156,0.28195) (0.63621,0.30452) (0.62919,0.32723)
    (0.62079,0.34945) (0.61126,0.37059) (0.60080,0.39013) (0.58956,0.40764)
    (0.57761,0.42276) (0.56499,0.43520) (0.55171,0.44470) (0.53775,0.45105)
    (0.52309,0.45410) (0.50771,0.45373) (0.49166,0.44994) (0.47502,0.44279)
    (0.45796,0.43253) (0.44076,0.41951) (0.42374,0.40422) (0.40730,0.38722)
    (0.39184,0.36909) (0.37776,0.35036) (0.36541,0.33150) (0.35512,0.31286)
    (0.34716,0.29473) (0.34179,0.27732) (0.33922,0.26076) (0.33966,0.24516)
    (0.34327,0.23058) (0.35016,0.21708) (0.36038,0.20470) (0.37385,0.19346)
    (0.39033,0.18342) (0.40945,0.17461) (0.43065,0.16708) (0.45327,0.16088)
    (0.47662,0.15608)
    };
  \draw[teal!75!black,line width=1.35pt,-{Latex[length=1.75mm,width=1.2mm]}] (0.64482,0.22103) -- (0.64495,0.26015);
  \draw[teal!75!black,line width=1.35pt,-{Latex[length=1.75mm,width=1.2mm]}] (0.35512,0.31286) -- (0.34179,0.27732);

  \fill[white] (C) circle (0.011);
  \fill[red!75!black] (C) circle (0.006);
  \node[anchor=west,fill=white,inner sep=0.5pt,font=\scriptsize]
       at ($(C)+(0.018,0.009)$) {$\bar P$};

  \fill[black!75] (P1) circle (0.008);
  \fill[black!75] (P2) circle (0.008);
  \fill[black!75] (P3) circle (0.008);
  \node[anchor=north east,font=\small] at ($(P1)+(-0.012,-0.014)$) {Token 1};
  \node[anchor=north west,font=\small] at ($(P2)+( 0.012,-0.014)$) {Token 2};
  \node[anchor=south,font=\small] at ($(P3)+(0,0.018)$) {Token 3};
\end{tikzpicture}

\caption{Phase portrait of the three-token example in
Section~4, with $c=7$. The unique equilibrium
$\bar P=(1/3,1/3,1/3)$ is a repelling focus.
The teal curve is a numerically computed attracting
periodic orbit of the limiting ODE. Trajectories
starting inside (orange) and outside (gray) the orbit
spiral toward it in a counterclockwise direction.
The shaded sectors indicate the most probable next
token: red for token~1, green for token~2, and blue
for token~3. The trajectories represent the limiting
deterministic dynamics, not individual stochastic
sample paths.
}
\label{fig:simplex-drift}
\end{figure}

Note that the limiting ODE in this construction is a cyclic three-strategy logit best-response dynamic, closely related to the rock–paper–scissors models studied by \cite{HommesOchea2012}.   Our contribution is to realize this dynamic within an explicit single-layer, attention-only transformer and establish the non-convergence of its empirical token frequencies.

Thus, we have a single-layer, NoPE, attention-only transformer that fails to converge almost surely at temperature $1$.  To extend this to any given temperature $\tau_0>0$, we multiply all of its
output embeddings by $\tau_0$.  Since output logits are divided by
$\tau_0$ before the softmax, the next token distribution
$q(P)$ is unchanged by this operation.  Thus the same ODE and 
the same non-convergence argument apply at temperature $\tau_0$. 
We obtain the following stronger statement.
\begin{theorem}
\label{thm:nonconvergence}
For every temperature $\tau_0>0$, there is a 
single-layer, NoPE, attention-only transformer with
no MLP whose empirical token distribution fails to converge almost surely when sampled at temperature $\tau_0$.
\end{theorem}
Recall Theorem~\ref{thm:multi-high-temp}, which fixes the transformer and then takes the temperature 
sufficiently large.  In contrast, in Theorem~\ref{thm:nonconvergence},
the temperature $\tau_0$ is fixed and 
the transformer is allowed to depend on $\tau_0$.  Thus every fixed transformer 
has a high temperature convergence regime, but there is no finite
temperature that guarantees convergence uniformly over all
transformers.

\section{Related Work}
\label{main:rel}

\paragraph{Dynamical and mean-field analyses of attention.}
Several recent works study self-attention as a dynamical or mean-field
system.  This includes work on clustering in self-attention
\citep{geshkovski2024emergence}, mean-field models with layer normalization
\citep{burger2025analysis}, meta-stable clustering
\citep{bruno2025emergence}, synchronization phenomena
\citep{polyanskiy2025synchronization}, and broader mean-field formulations
of transformer dynamics \citep{rigollet2025meanfield,tembine2024mean}.
Other work gives exact mean-field analyses in simplified attention models
\citep{marion2025understanding} and studies mean-field descriptions of
in-context feature learning \citep{kim2024transformers}.  These papers
primarily analyze the evolution of representations across layers, training,
or an idealized population.  Our setting instead follows a fixed
autoregressive model forward in generation time, where the empirical
history itself changes after every generated token.

There are also other mathematical views of attention.
\citet{katharopoulos2020transformers} show that linearized autoregressive
attention admits a recurrent formulation, \citet{ramsauer2021hopfield}
relate softmax attention to modern Hopfield networks and associative memory,
and \citet{dong2021attention} study rank collapse under repeated pure
self-attention.  Our use of stochastic approximation follows
\citet{benaim1999dynamics} to compare a slowly changing
stochastic process with an ODE.

\paragraph{Related dynamical analyses.}

Other works include symmetry breaking in attention
~\citep{silverstein2026symmetry},
synchronization-inspired transformer
architectures~\citep{liu2026krause}, and residual-aware cumulative
attention dynamics~\citep{Herasimchyk26}. These works study
related dynamical phenomena, but not the infinite-generation limit
we consider.

\paragraph{Transformers and positional information.}

\citet{vaswani2017attention} use absolute sinusoidal position encodings while later work use relative position representations, e.g., directly into self-attention~\citep{shaw2018relative},  combining a recurrent memory across segments \citep{dai2019transformerxl}, or adding a distance-dependent linear bias to the attention scores
\citep{press2022alibi}.  RoPE rotates query and key vectors as a function of position so that their inner product depends on relative position \citep{su2021roformer}.

The choice of positional representation is closely related to length
generalization.  \citet{kazemnejad2023positional} compare several positional
schemes, including RoPE, ALiBi, and NoPE, and observe good length
generalization for NoPE on many tasks. 

This phenomenon has also been studied theoretically for NoPE
\citep{wang2024length}, and more broadly for length generalization
in causal transformers \citep{huang2025formal}.  Randomized positional encodings
have also been used in this context \citep{ruoss2023randomized}.

\paragraph{Length extrapolation and RoPE.}
A different line of research asks how a model trained at one context length can be
used at a larger one.  \citet{sun2023length} study architectural choices for
length extrapolation.  Position interpolation rescales position indices for
pretrained RoPE models \citep{chen2023position}, YaRN modifies RoPE scaling
to extend context windows efficiently \citep{peng2024yarn}, and LongRoPE uses
non-uniform interpolation and progressive extension to reach longer
contexts \citep{ding2024longrope}.  Other work studies limitations and
variants of RoPE itself, including the relation between the RoPE base and
usable context length \citep{xu2024base}, the distribution of information
across RoPE frequencies \citep{probing2026rope}, and wavelet-based rotary
encodings \citep{reid2026wire}.  These works are concerned with
extrapolating a trained model to longer inputs.  We, on the other hand, study the limiting behavior of a fixed model.

\paragraph{Long-context attention and memory.}
Another line of work changes the attention or memory mechanism itself.
Transformer-XL reuses hidden states from previous segments
\citep{dai2019transformerxl}, and the Compressive Transformer stores older
states in compressed form \citep{rae2020compressive}.  Longformer combines
local windowed attention with selected global attention
\citep{beltagy2020longformer}, while BigBird uses a sparse mixture of local,
random, and global connections \citep{zaheer2020bigbird}.  Memorizing
Transformers augment the model with retrieval from a memory of past
key--value representations \citep{wu2022memorizing}.  StreamingLLM retains
a small set of initial ``attention sink'' tokens together with a recent
window \citep{xiao2024streamingllm}, while Infini-attention combines local
attention with a bounded compressive memory \citep{munkhdalai2024infini}.
In contrast, we keep ordinary global attention and study the statistics of
its growing KV cache.

\section{Conclusion}

We studied the long-run behavior of transformers 
through the slowly changing empirical state generated by their own
outputs.
In both the single-layer and multi-layer settings, the
stochastic process tracks an ODE on an appropriate time
scale.  For every fixed transformer in our model, sufficiently high
temperature makes this ODE contractive and forces convergence to a
unique limiting state.  This threshold is model-dependent:
at every fixed finite temperature, some transformer can have a limiting ODE that does not converge.  Thus temperature controls long-run
stability, but there is no universal finite temperature that guarantees
convergence for all transformers.

\subsection*{AI use statement}

In this work, we used generative AI tools only to \emph{formulate mathematical claims; provide critical ingredients for proving mathematical claims; assist in the writing of proofs}.
We have not used generative AI tools to help develop theoretical models or conceptual frameworks; propose or refine hypotheses; interpret results; generate synthetic data sets;  design or provide feedback on research methodology or experiments; implement methods; assist with translation; clean and reformat datasets; support qualitative and thematic data analysis are not applicable to this work. Additionally, we used generative AI tools to \emph{draft parts of a research paper; identify relevant literature}.  All LLM-assisted proofs were checked and fully rewritten by the authors.  We take responsibility for the final content of this work, including text, claims, or artifacts produced with the aid of generative AI.

\bibliography{refs}
\bibliographystyle{plainnat}

\end{document}